\documentclass{article}

\PassOptionsToPackage{numbers, compress}{natbib}

\usepackage[preprint]{neurips_2020}

\usepackage[utf8]{inputenc} 
\usepackage[T1]{fontenc}    
\usepackage{hyperref}       
\usepackage{url}            
\usepackage{booktabs}       
\usepackage{amsfonts}       
\usepackage{nicefrac}       
\usepackage{microtype}      
\usepackage{todonotes}
 \def\etal{\textit{et al.}}
\title{Recent Advancements in Self-Supervised Paradigms for Visual Feature Representation}

\author{%
  Mrinal Anand \\
  Department of Computer Science\\
  Indian Institute of Technology, Gandhinagar\\
  \texttt{mrinal.anand@iitgn.ac.in} \\
   \And
   Aditya Garg \\
   Department of Computer Science\\
  Indian Institute of Technology, Gandhinagar \\
  \texttt{aditya.garg@alumni.iitgn.ac.in} \\
}

\begin{document}
\maketitle

\begin{abstract}
We witnessed a massive growth in the supervised learning paradigm in the past decade. Supervised learning requires a large amount of labeled data to reach state-of-the-art performance. However, labeling the samples requires a lot of human annotation. To avoid the cost of labeling data, self-supervised methods were proposed to make use of largely available unlabeled data. This study conducts a comprehensive and insightful survey and analysis of recent developments in self-supervised paradigm for feature representation. In this paper, we investigate the factors affecting the usefulness of self-supervision under different settings. We present some of the key insights concerning two different approaches in self-supervision, generative and contrastive methods. We also investigate the limitations of supervised adversarial training and how self-supervision can help overcome those limitations. We then move on to discuss the limitations and challenges in effectively using self-supervision for visual tasks. Finally, we highlight some open problems and point out future research directions.
\end{abstract}

\section{Introduction}
Supervised learning is a popular choice to solve many machine learning problems. A large impact has been observed in the field of computer vision under the supervised setting, i.e. learning from labeled data. Deep Neural Networks have shown remarkable results when used on problems such as image classification~\cite{he2015deep,szegedy2016inceptionv4}, object detection~\cite{beery2020context}, image generation~\cite{Infogan} and semantic segmentation~\cite{chen2018encoderdecoder,howard2019searching}. Under supervised settings, a model is trained on the human-annotated labels. Supervised learning encompasses models where for every corresponding input, we are given a target label, and the model is then trained to learn that label.

These remarkable performances of supervised learning algorithms result from the large number of human annotations needed to label the sophisticated datasets. However, collecting and manual labeling of large datasets are expensive and time-consuming. ImageNet~\cite{imagenet} has around 14 Million images, and each image might fall in more than one out of 22000 classes. This example showcases the expense incurred to collect large, curated datasets. This has motivated the researchers to explore the domain of unsupervised learning and to learn from the vast amount of unlabeled data available.

Unsupervised learning refers to those algorithms that use the input data to infer patterns without any explicit supervision. Self-supervised learning is a subset of unsupervised learning where labels are generated automatically.
In this work, we study the recent advancements in self-supervised learning that have been proposed to learn robust features representation from the unlabelled data. It is also seen as a way to use a large amount of untapped data available on the internet. This approach has been around for some time, but the recent advancements have shown remarkable results that are comparable or better to those shown by supervised approaches.

The idea for self-supervision comes from contextual embeddings used in natural language processing. The idea here is to map every word to a feature vector such that a word can be predicted if the context(a few words before or after) is given. In essence, context prediction is just a \textit{pretext} for the network to learn effective word embeddings. A similar method is used in a self-supervised setting where tasks are performed under the pretext of learning representations. Through this work, we aim to showcase a trend that has been followed in the recent advancements in the field of self-supervised learning and give insights that could be helpful for machine learning practitioners.

\noindent \textit{Outline of the paper:} The entire paper is organized as follows. Section \ref{Pretext} narrates the different pretext tasks used in self-supervised learning methodology. Section \ref{paradigm} describes the generative and contrastive learning paradigms. Section \ref{Pretraining} presents the important factors affecting the learned representation through pre-training. Section \ref{Robustness} explores the adversarial robustness in the context of self-supervision. Finally, Section \ref{discussion} discusses the limitations in the current methodology and present the future directions for the self-supervised paradigm.

\section{Pretext Tasks for Self-Supervision}\label{Pretext}
\begin{figure}[!h]
    \centering
    \includegraphics[scale=0.5]{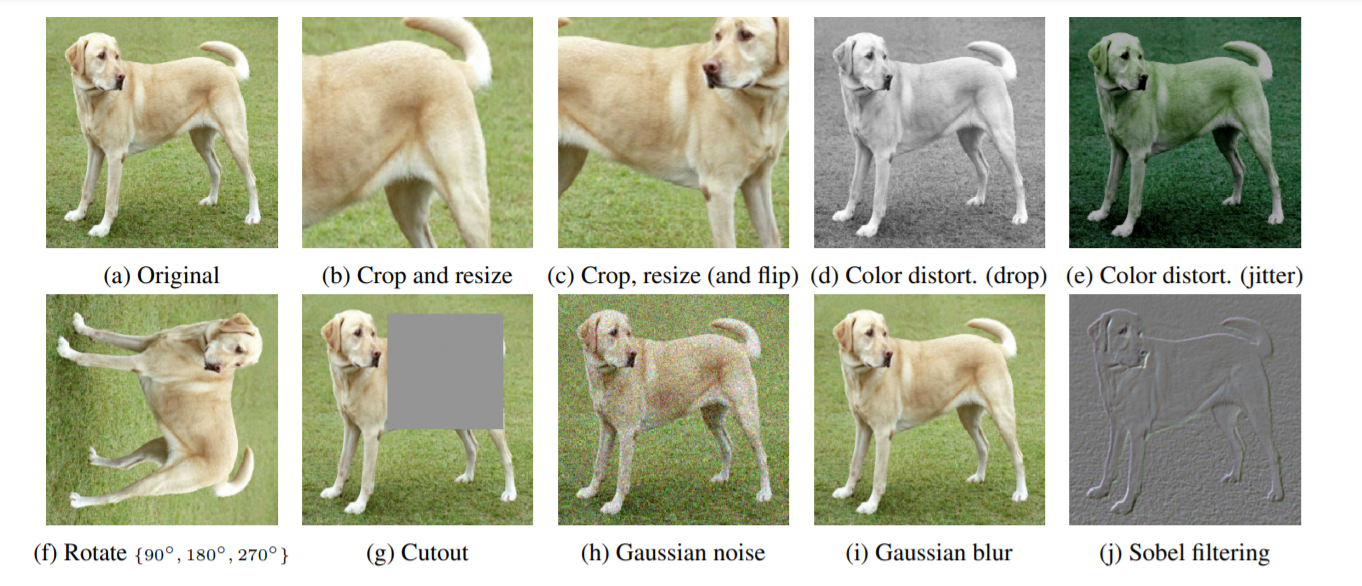}
    \caption{Different augmentation techniques used as pretext tasks (~image source~\cite{chen2020simple}~).} 
    \label{augmentations}
\end{figure}

A pretext task is a proxy task for the network to solve with the goal of learning feature representations that can be easily transferable to specific downstream tasks. It is ensured that pretext tasks are as hard as possible, and also that original data is in no way destroyed. Downstream tasks are the primary vision tasks for which the self-supervised model is being trained. The performance of the model on these tasks determines the success of the model. Downstream tasks can be anything ranging from object detection, object classification, or image segmentation. Popular augmentations that are being used extensively in self-supervised models are shown in Figure \ref{augmentations}.
Some of the popular pretext tasks that are used in self-supervised learning:
\begin{enumerate}
    \item \textbf{Exemplar CNN}~\cite{dosovitskiy2015discriminative} -
    N patches are sampled from every image from different locations in the image and scaled randomly. Patches should be picked from parts where the gradient is thicker as these parts usually carry the different objects. Every patch is augmented in a variety of different ways, and the pretext task would be to differentiate between patches of different images. The feature representation learned performs well in object classification and descriptor matching.
    \item \textbf{Rotation}~\cite{gidaris2018unsupervised} - An image is rotated by a multiple of 90°. Then the model is trained to classify the rotation into four classes, namely [0, 90, 180, 270]. This is also an effective way to force the model to learn high-level features. They demonstrate that rotation transformation that was applied to an image will require the model to understand the concept of the location of an object in the image, its type, and its pose.
    \item \textbf{Relative Patches}~\cite{doersch2015unsupervised} - This task extracts multiple patches from an image and asks for the relation between them. One patch is randomly sampled, and another patch is sampled from the eight neighboring patches. Now the relative position is asked between them. To prevent low level features from slipping, additional noise is added. This task has been found to be useful for both object detection and visual data mining. This has also been adapted into a \textbf{Jigsaw puzzle game}~\cite{noroozi2016unsupervised} task for the model to learn. This jigsaw puzzle has been seen to be useful for object detection and classification.
    \item \textbf{Colorization}~\cite{zhang2016colorful} - A model is trained to appropriately color a grayscale image. It is surprisingly useful as a pre-training task for diverse downstream tasks like object classification, detection, and segmentation. However, evaluating the performance requires a manual human intervention which can be very time-consuming.
\end{enumerate}

\section{Two Paradigms for Self-Supervision}
\label{paradigm}
In this section, we will discuss the two main approaches in statistical modeling. The first one being generative modeling and other one is discriminative approach. In the generative approach models the conditional probability of the input given the target while the discriminative models the conditional probability of the target given the input. 

\subsection{Generative Learning}
Majority of the work in generative image learning involves training the model to reconstruct the original data in great detail. Their goal this way is to also learn semantically meaningful representation. These generative modeling techniques are very powerful and have been found to represent the data very well. 

\textbf{Autoregressive Models:} An effective line of approach in generative learning is to model it as a product of conditional distributions. Effectively, we then have to predict the future instances given the past instances. Recent models that have shown some levels of success are \textit{PixelCNN}~\cite{van2016conditional} and \textit{PixelRNN}~\cite{oord2016pixel}. These models are known to be computationally expensive because they try to model the extremely complex relationship between pixels. 

\textbf{Autoencoding Models:} Traditional autoencoding models encode the input and then decode it to give an output, which is as similar to the input as possible. The model is forced to learn a function that encodes the input into a representation, and then this representation is later decoded back to the original image. The task here is to learn a function that effectively captures the semantic features in the data. This function is then used in downstream tasks. Autoencoders cannot handle complex images easily because it tries to capture as much information as possible but not all the relevant information. This causes the decoded image to be degraded. There also presents the risk of overfitting when this kind of autoencoding model is used for feature representation.

\hspace{0.5cm}\textit{Denoising Autoencoder}~\cite{vincent2008extracting} was introduced to rectify the overfitting in traditional autoencoding models and also to improve the robustness of the model. It introduces some random noise in the original image and trains the traditional autoencoder to reproduce the original image. An interesting fact was noticed during experiments for this model. As the noise is increased, the model tends to learn more about larger structures that span across multiple input dimensions and thus capture more important features. This makes it a good pre-training task. The problem with this kind of noise addition is that it still is pretty low level and does not force the model to learn a thorough semantic understanding of the image.

\hspace{0.5cm}\textit{Context Encoder}~\cite{pathak2016context} was proposed to learn a more semantic meaningful representation. Instead of adding a low-level noise, we remove a patch from an image and force the model to fill the patch and output an image similar to the original image. The paper provides interesting intuition regarding the use of loss in their model. They demonstrate that simple L2 loss between the two images results in the model learning to paint a blurry picture. They hypothesize that this might be because it is 'safer' for the model to predict the mean of the distribution. They show that having both adversarial loss and L2 loss makes the inpainting more realistic. Zhang \etal~\cite{zhang2016colorful} demonstrated that context encoder does not perform well on large-scale semantic representation learning benchmarks. This might be because the loss function is not able to capture inpainting quality properly, or it could be an easy task that can be solved without much semantically meaningful information and just by using the local information. Though, it has been seen that context, encoder works nicely for the object detection task.

\hspace{0.5cm}\textit{Split Brain Autoencoder}~\cite{zhang2017split} is another kind of autoencoder where the color channels of an image are randomly divided into two mutually exclusive subsets, and then the network is divided into two disjoint sub-networks where each subnetwork is then trained to predict the other subset of data. It has been seen to have an excellent performance for image classification.

\hspace{0.5cm}\textit{Variational Autoencoders}~\cite{kingma2013auto} has spawned many variations, and these variations have been successful in modeling data. It works differently from the other variants of autoencoder because it maps the data into a distribution rather than a vector representation. So, instead of encoding the data into a vector, we encode it into distribution and then use that distribution to decode the original input data. Some popular variants are \textit{VQ-VAE}~\cite{van2017neural} and \textit{VQ-VAE2}~\cite{razavi2019generating}. VQ-VAE has shown that it can be used to model the long-term dependencies, and thus many important features can also be captured.

\textbf{Generative Adversarial Models:} Popularly known as GANs~\cite{goodfellow2014generative} have come up as a powerful tool to learn models that stem from very complex distribution. It consists of two different networks, the generator, and the discriminator. The generator synthesize image in order to fool the discriminator in believing that the synthesized image is real. Radford \etal~\cite{radford2015unsupervised} has shown that GANs are able to capture powerful semantic information, which can be very useful in self-supervised pretraining. To use these features for representation learning, Donahue \etal ~\cite{donahue2016adversarial} came up with a novel Generative architecture \textit{BiGAN}. Along with the normal discriminator and generator, this model also includes another network called the encoder. This encoder turns the real images into latent representations. The task here is then similar to that of an autoencoder where the generator and the encoder have to be inverse of each other. BiGAN has been shown to be effective in learning visual representations. This can be classified as a  self-supervised method because the discriminator learns to assign labels such as 'fake' or 'real' to images and thus self-supervised learning is involved in all kinds of GANs.

\hspace{0.5cm}\textit{BigBiGAN}~\cite{donahue2019large}, a modification to BiGANs, replaces the BiGAN generator with \textit{BigGAN}~\cite{brock2018large}'s generator. BigGAN is a more modern GAN architecture that has proved more capable of capturing useful semantic features. 

\subsection{Contrastive Learning}
Contrastive learning has come up recently as an alternative to generative modelling. The objective here is to have the model learn to compare between similar and dissimilar examples. The objective function used here is the Noise Contrastive Estimation (NCE)~\cite{gutmann2010noise}. The NCE objective is formulated as follows:
\begin{equation}
    \mathop{\mathbb{E}[-\log\frac{f(x, x^+)}{f(x, x^+) + \sum_{k=1}^{K}f(x, x_k^-)}]}
\label{closs}
\end{equation}
Formulation of the NCE loss function (\ref{closs}). $x^+$ is the augmented version of x, and $x_k^-$ are dissimilar to x. K is the number of negative samples we want to compare our sample x to.

In the recent past, we have witnessed a large amount of models that uses this contrastive loss in different settings. Models that are based on maximizing mutual information have been around for a long time in unsupervised learning. Mutual information (MI)~\cite{belghazi2018mine,becker1996mutual,bell1995information} is a measure of the dependence between two signals. \cite{hjelm2019learning} defines mutual information between two variables A and B as the decrease in uncertainty of A if we have prior knowledge about B. The task is modeled around learning a representation while also maximizing this MI between both the inputs and representation. MI has always been difficult to compute, but the recent breakthrough in techniques to efficiently approximate MI has made these models possible.

Recent work like \textit{Contrastive Predictive Coding}~\cite{oord2018representation} have moved away with maximizing MI~\cite{hjelm2018learning} between the input and the learned representations but instead have tried to maximize the lower bound on MI~\cite{belghazi2018mine}. They theoretically show how the lower bound on MI increases as the Contrastive Loss decreases. In CPC for visual tasks, the role of the pretext task is to predict the representations of a patch somewhere below another patch from other patches. The task here is similar to the one performed in an autoregressive model. 
\begin{figure}[!h]
    \centering
    \includegraphics[scale=0.5]{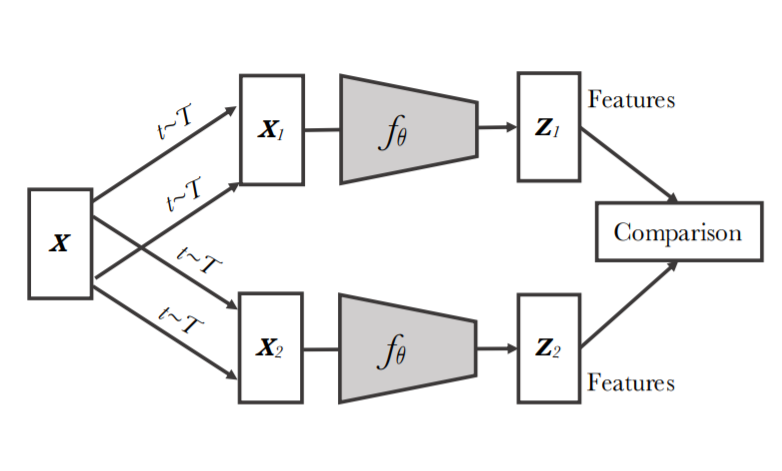}
    \caption{Constrastive Instance Learning. X is an image from the dataset. T is the set of image transformations. $\theta$ is the parameters for the encoder. At the end we compare with the negative samples (~image source~\cite{caron2020unsupervised}~).} 
    \label{Constrastive}
\end{figure}

The early success of models that maximize the mutual information was challenged by Tschannen \etal~\cite{tschannen2019mutual}, where they argued with credible evidence that maximizing mutual independence was not the main reason for the success of the early models. They also empirically show that MI maximization and representation quality are not co-related. They believe that the success lies in the encoder architecture and negative sampling.

\hspace{0.5cm}\textit{Instance level discrimination}~\cite{wu2018unsupervised} is a pretext task which forms the basis of the many successful recent models. Each image is encoded into a feature representation, and then they are spread across the feature space so that similar samples are not as far as dissimilar samples. The distance metric would be the L2 distance between the representations.
Based on CPC and Instance Discrimination, another model called \textit{Contrastive MultiView Coding} (CMC)~\cite{tian2019contrastive} was proposed. Instead of using spatial prediction as the pretext task, this model learns to create representations by comparing against multiple augmentations of an image. As the number of views increases, the performance becomes better but at the cost of computational power. Similar to CMC, \textit{Augmented Multiscale Deep InfoMax} (AMDIM)~\cite{bachman2019learning} was proposed. The major difference between AMDIM and CMC is that AMDIM maps two augmented versions of the original image to as well as their intermediate representation to the similar space while CMC works on matching image channels in a contrastive fashion.  

It has been observed that contrastive learning seems to benefit from a large number of negative samples. This observation, combined with the instance level discrimination, inspired MoCo or \textit{Momentum Contrast}~\cite{he2020momentum}. It has a queue that contains the recently encoded batches, which are used as the negative samples. The samples in the queue are known as keys, and the query is the sample that is encoded and then compared to the keys. An important information to note about MoCo is that it does not run backpropagation on all the samples in the queue.
Instead, it backpropagates on the query encoder, and a momentum update is run on the parameters of the key encoder. It performs instance-level discrimination but brings in more negative samples, and so is a self-supervised learning technique.

\hspace{0.5cm}\textit{Pretext Invariant Representation Learning} or PIRL~\cite{misra2020self} tries to learn representations that are invariant to the augmentations applied on them. It transforms an image and then encourages the representations of the augmented image and original image to be invariant. The pretext task here is to learn that augmented representations of the same image are invariant. \textit{SimCLR}~\cite{chen2020simple} was introduced later, which trained the model in an end to end manner. It is similar in theory to PIRL, but it generalizes the approach and adds a non-linear projection on top of the representation layer to boost the model's accuracy.
Batches of larger size, which considerably affect the computation cost, have to be used but are necessary for the model's success.
A generalized version of these models is Figure \ref{Constrastive}. 

The SimCLR shed light on the importance of the choice of data augmentation operations in the success of a model. They highlight the effectiveness of the composition of random cropping and random color distortion augmentations in their model. They have performed further analysis to show that this non-linear projection layer actually helps maintain important information that otherwise would have been removed by the contrastive loss. These studies helped MoCo, and they weeded out the deficiencies in their model. \textit{MoCo v2}~\cite{chen2020improved} is the enhanced version of MoCo and utilizes some of the techniques from SimCLR to provide a model that is computationally less expensive and also gives better performance than SimCLR.

\hspace{0.5cm}Clustering-based methods such as \textit{DeepCluster}~\cite{caron2018deep} uses a simple method such as k-means and an encoder to learn representations for the visual features. DeepCluster uses k-means algorithm, where it first assigns every feature vector to its nearest center. The parameters of the network are then optimized on the basis of the assigned labels. These two steps are then performed repeatedly.

\hspace{0.5cm}\textit{Local Aggregation}~\cite{zhuang2019local}, an improvement over DeepCluster, scraps the formation of hard clusters, and it forms soft clusters, which has a much lower computational cost compared to DeepCluster. It optimizes a soft clustering metric which is computationally more efficient than the cross-entropy function that needs to be recalculated at every step. These changes have led to better performances on different benchmarks and also better computational efficiency. 

\hspace{0.5cm}\textit{SwAV}~\cite{caron2020unsupervised} is a recent model that proposes the use of clustering to learn effective representations. The normal contrastive model uses pairwise comparisons between data, but this model here discriminates between groups of images with similar features. Unlike contrastive methods, this model works well even with small batches. Thus it is computationally very inexpensive. The paper also empirically demonstrates the potential of cluster-based approaches for self-supervised learning. It modifies DeepCluster~\cite{caron2018deep} to give results superior to even SimClR~\cite{chen2020simple}. Their own model SwAV gives results on par with DeepCluster v2, but SwAV has its own advantages. Not only is it an online model, but it also scales very nicely. It is very suitable for large datasets.

Table \ref{acc} compares the accuracy of some recent models on the ImageNet~\cite{imagenet} dataset. As shown in the table some of contrastive methods like \textit{SwAV} and \textit{MoCo v2} achieves near the supervised accuracy.
\begin{table}
\centering
 \begin{tabular}{c c} 
\toprule
 \textbf{Method} & \textbf{Accuracy} \\ [0.5ex] 
 \toprule
 
 Supervised & \textbf{76.5}\% \\
 
 Local Aggregation & 58.8\%\\
 
 MoCo v2. & 71.1\%\\
 
 MoCo & 60.6\%\\

 PIRL & 63.6\%  \\
 
 CPC v2. & 63.8\% \\
 
 SimCLR & 70.0\%  \\
 
 BigBiGAN & 56.6\% \\
 
 SwAV & 75.3\%\\ 
 \toprule
\end{tabular}
\caption{Table to compare \textbf{Top 1} accuracy on \textbf{ImageNet}~\cite{imagenet} dataset between popular self-supervised models. All these results are for models using ResNet-50~\cite{he2015deep} architecture with \textbf{24 Million} Parameters. Note that we use frozen weights from the models to measure performance.}
\label{acc}
\end{table}

\section{Self-Supervised Pretraining}\label{Pretraining}
Pretraining has shown a significant performance boost on visual tasks~\cite{donahue2013decaf}. Pretraining a model with large datasets like ImageNet~\cite{imagenet} and MS COCO~\cite{mscoco} helps the model to converge faster. Moreover, it also helps in transferring the learned representation across different visual tasks. The aim of self-supervised pretraining is to learn a robust feature representation from unlabelled data, such that fewer and fewer labeled samples are required to train a model to its maximum capacity. Generally, deep models largely benefit from pretraining when labeled data are scarce. However, practical usage of self-supervised pretraining depends on several other aspects also, Newell \etal~\cite{newell} presents four different factors that affect the performance of pretraining-
\begin{itemize}
    \item \textbf{Data:} The feature learned through self-supervision directly depends on the complexity of the dataset. The complexity of the dataset is linked to the diversity of the samples and the number of labeled samples to saturate the performance on a task.
    \item \textbf{Model:} The maximum accuracy achieved depends upon the backbone architecture of the model. Model having large parameters generally have high expressivity power. Hence it can better generalize on unseen examples.
    \item \textbf{Pretext Task and Proxy loss:} The pre-text task has a strong influence on the type features learned, and therefore the choice of pre-text task can affect the performance of the downstream task. Usually, the pre-text task like Rotation~\cite{gidaris2018unsupervised} and Jigsaw Puzzle~\cite{noroozi2016unsupervised} learns global representation, and task like Colorization~\cite{zhang2016colorful} learn more semantic/boundary level representation. The choice between contrastive or pixel-wise loss also affects the type of representation learned through self-supervision.
    \item \textbf{Downstream Task:} The different downstream tasks require different kinds of supervised signals and can largely benefit if a suitable pre-text task is used for training. Broadly any downstream task can be either a semantic heavy or geometric heavy task. Another significant distinction between these vision tasks is the requirement of dense prediction along with the image space, i.e., per-pixel wise prediction or single prediction per sample. Different pre-text tasks have a different affects on the performance of the model. However, self-supervised learning cannot be used for all downstream tasks since it relies on automatic label generation.
\end{itemize}

Now we highlight some of the major findings related to the choice of the correct pre-text task for a given downstream task and other associated insights -  

\textbf{Number of Labeled Samples:} Self-supervised pretraining has a significant impact on the performance when the number of labeled samples is small. The gain in the performance gets negligible as the number of labeled samples increases. This observation is unaffected by the choice of pretext task or the choice of downstream task.

\textbf{Affect on Downstream Task:} It has been observed that CMC based pretraining performs better than other pretext tasks like AMDIM, Rotation, and VAEs for object detection and pose estimation. On the contrary, rotation and AMDIM perform better on low-level tasks like segmentation and depth estimation~\cite{newell}. One of the possible reason could be CMC applies a contrastive loss to global features while AMDIM and Rotation motivates to match similar low-level features to similar space.

\textbf{Model Selection:} Popular choices of model for pre-training are ResNet~\cite{he2015deep}, VGG~\cite{simonyan2015deep} and AlexNet~\cite{alexnet}. The choice of model largely depends on the type of downstream task. For capturing more fine-grained information, VGG based model having a small kernel size and small stride is preferred over AlexNet based model, which have relatively large kernels and stride. Using large kernels imposes too many parameters in the model to tune, and a large convolution stride can ignore some fine-grained features in the lower layers.

\textbf{Affect of freezing the weights:} Freezing the weights restricts the overall expressive power of the deep neural networks. It has been empirically studied that performance suffers from freezing the layer weights~\cite{newell}, but on the contrary, it is advisable to freeze the model layers if the number of labeled samples is very low; otherwise, it can lead to overfitting. 

\section{Adversarial Robustness through Self-Supervision}\label{Robustness}
Adversarial Attacks are the perturbed inputs especially designed to fool neural network~\cite{Goodfellow2015ExplainingAH}, and Adversarial Training has shown great results in providing Adversarial robustness to a deep model. 
However, performing supervised adversarial training in model space, i.e., optimizing the model parameters, has several drawbacks on its own.
\begin{itemize}
    \item \textbf{Accuracy Drop:} Model training using adversarial loss is generally accompanied by a significant drop in accuracy.
    \item \textbf{Label Leakage:} Label Leakage happens when the performance of the model is better on adversarial images rather than clean images\cite{Kurakin2017AdversarialML}. This happens because the true label information leaks on the adversarial examples. 
    \item \textbf{Task Dependency:} A model trained adversarially for object detection cannot provide the same adversarial robustness to classification models.  
\end{itemize}

A different line of research focuses on processing the input as an adversarial defense~\cite{guo2018countering,zeng2020data}. Guo \etal~\cite{guo2018countering} studies bit-depth reduction, JPEG compression, total variance minimization, and image quilting as the perturbation in the input space. Athalye \etal~\cite{Athalye2018ObfuscatedGG} showed that these defense suffers from obfuscated gradients and can easily be broken under white-box settings.

\begin{figure*}[!t]
    \centering \includegraphics[width=\linewidth]{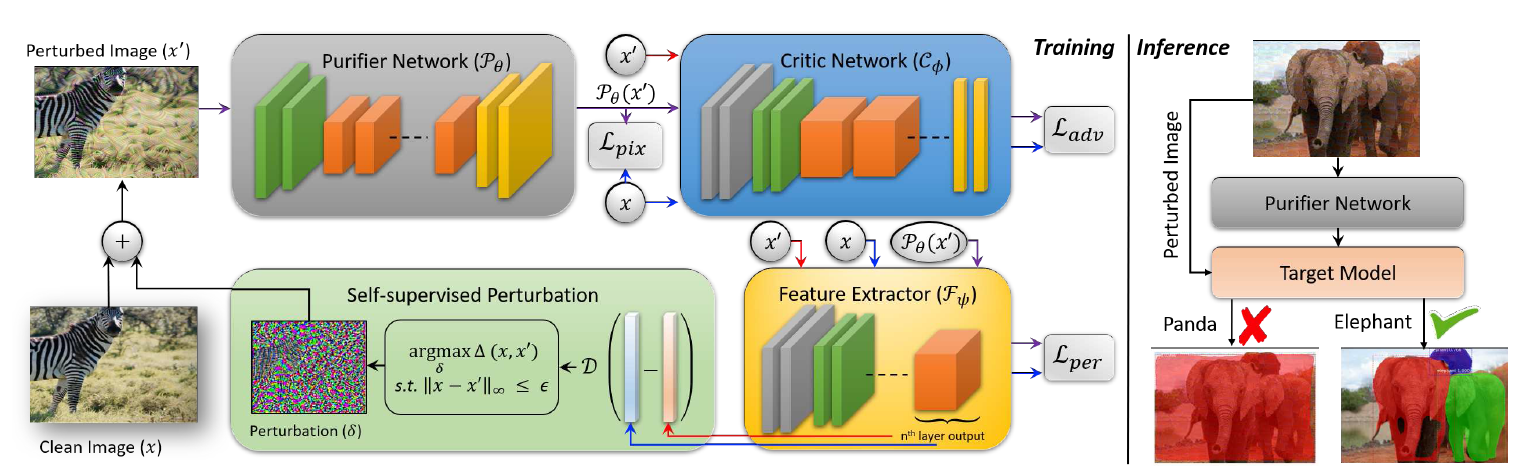}
\caption{Neural Representation Purifier Network that generates noisy free images from perturbed input images (~image source~\cite{naseer}~).}
\label{fig:purify}
\vspace{-0.13in}
\end{figure*}

Naseer \etal~\cite{naseer} proposed a Neural Representation Purifier that works on input space. The proposed technique is formulated as a min-max game that learns an optimal input processing function that enhances robustness in a self-supervised setting. Figure \ref{fig:purify} depicts the whole pipeline in which a purifier network generates a clean image from a given perturbed image and a self-supervised perturbation (SSP) network that generates the perturbation using deep perceptual features of the input image. At the time of inference, we simply add this adversarially trained purifier network before the target model, as shown in Figure \ref{fig:purify}. 

We want to highlight a few points for the readers that using adversarial training in a self-supervised manner by default eliminates the problem of label leakage as no true labels are involved while training. Moreover, working in input space rather than model space removes the problem of accuracy drop and task dependency as original model parameters remain unaffected.

\section{Discussion and Future Directions}
\label{discussion}
Through diverse experimentation, Kolesnikov \etal~\cite{ssv} showed that neither the architectures are coherent with the pretext task nor the self-supervised pretext task is consistent with the architectures in terms of performance. One possible future direction is to develop such a pretext task that is practically invariant to the downstream task, or in other words, it learns domain invariant representation that can easily be transferred to different vision tasks.

There is no concept of transfer across datasets in visual feature learning. Models like BERT~\cite{devlin2018bert} have been trained on large text corpora, and these models can be used for many other applications on different datasets after fine-tuning and are shown to give excellent results.

As demonstrated by Song \etal~\cite{song2019understanding}, alternative measurements of information have to be looked for as Mutual Information estimation has been found to be challenging, and there is no satisfactory method that optimizes Mutual information. Mutual Information Estimators that are being used have lots of limitations and often lead to exponentially growing variance.

All recent state-of-the-art contrastive paradigms use a similar technique, which is to push representations of different views of the same image closer together and for different images as far as possible. The view mentioned here is the same one mentioned for Contrastive Multiview Coding~\cite{tian2019contrastive}.
Tian \etal~\cite{tian2020makes} have presented a paper that shows how critical view selection is. They discuss how important it is to select views that have a balanced invariance, that captures just enough information to label the images. It is also discussed how each downstream task needs a different set of optimal views. They also present a model that outperforms many of the state of the art models.
This shows that there is still a distance to go in terms of selecting optimal views for downstream tasks.

There is a lot of ground to cover in regards to the theoretical understanding of the state of the art models. We still do not have a very clear reason for the success of Mutual Information based models~\cite{tschannen2019mutual}.

Self supervised learning algorithms have shown promising results in vision so far and it could be the next revolution in the field of visual feature learning.

\section{Conclusion}
We discussed compelling results that have been achieved by the recent models in the self-supervised domain. These models have bridged the gap between unsupervised and supervised learning considerably.
In this work, we first discussed the diverse pretext task used for self-supervision and investigated how different pretext tasks correlate to different downstream tasks. We then discussed some of the important works which follow the two different methodologies in self-supervised learning, i.e., generative and contrastive learning. The section also gives insights related to their methodologies. We further present the factor involved in pretraining a model under self-supervised settings and investigate some of the competing choices that machine learning practitioner encounters while using self-supervision. We then briefly touch upon the challenges in supervised adversarial learning and how self-supervision helps tackle those problems. Finally, we end with a discussion on limitations in current methodology and point out some future research directions.

\bibliographystyle{plain}
\bibliography{refs}

\end{document}